\documentclass[sigconf, nonacm]{acmart}
\usepackage{algorithm}
\usepackage{algpseudocode}
\usepackage{float}
\usepackage{pifont}
\usepackage{xcolor}
\usepackage{tcolorbox}
\usepackage{alltt}
\usepackage{array, makecell}
\usepackage{multirow}
\usepackage{csquotes}
\usepackage{listings}
\usepackage{xcolor}
\usepackage{amsmath}

\newcommand{\cleanagent}{\textsc{CleanAgent\ }}
\theoremstyle{plain}

\definecolor{codegreen}{rgb}{0,0.6,0}
\definecolor{codegray}{rgb}{0.5,0.5,0.5}
\definecolor{codepurple}{rgb}{0.58,0,0.82}
\definecolor{backcolour}{rgb}{0.95,0.95,0.92}

\lstdefinestyle{mystyle}{
    backgroundcolor=\color{backcolour},   
    commentstyle=\color{codegreen},
    keywordstyle=\color{magenta},
    numberstyle=\tiny\color{codegray},
    stringstyle=\color{codepurple},
    basicstyle=\ttfamily\footnotesize,
    breakatwhitespace=false,         
    breaklines=true,                 
    captionpos=b,                    
    keepspaces=true,                 
    numbers=left,                    
    numbersep=5pt,                  
    showspaces=false,                
    showstringspaces=false,
    showtabs=false,                  
    tabsize=2
}
\begin{document}
\title{CleanAgent: Automating Data Standardization with LLM-based Agents}
\author{{Danrui Qi, Zhengjie Miao, Jiannan Wang}}
 
\affiliation{%
  \institution{\hspace{0em} Simon Fraser University \\
  {\{dqi, zhengjie, jnwang\}@sfu.ca}}
  \country{}
}

\begin{abstract}
Data standardization is a crucial part of the data science life cycle. While tools like Pandas offer robust functionalities, their complexity and the manual effort required for customizing code to diverse column types pose significant challenges. Although large language models (LLMs) like ChatGPT have shown promise in automating this process through natural language understanding and code generation, it still demands expert-level programming knowledge and continuous interaction for prompt refinement.
To solve these challenges, our key idea is to propose a Python library with declarative, unified APIs for standardizing different column types, simplifying the LLM's code generation with concise API calls. We first propose \texttt{Dataprep.Clean}, a component of the Dataprep Python Library, significantly reduces the coding complexity by enabling the standardization of specific column types with a single line of code. Then, we introduce the \cleanagent framework integrating \texttt{Dataprep.Clean} and LLM-based agents to automate the data standardization process. With \cleanagent, data scientists only need to provide their requirements once, allowing for a hands-free process. 

\end{abstract}

\maketitle

\vspace{-.3em}
\section{Introduction}

Data standardization, which is pivotal in the realm of data science, aims to transform heterogeneous data formats within a single column into a unified data format. This crucial data preprocessing step is essential for enabling effective data integration, data analysis, and decision-making. 

\begin{figure}[t] 
\centering 
\includegraphics[width=1.0\linewidth]{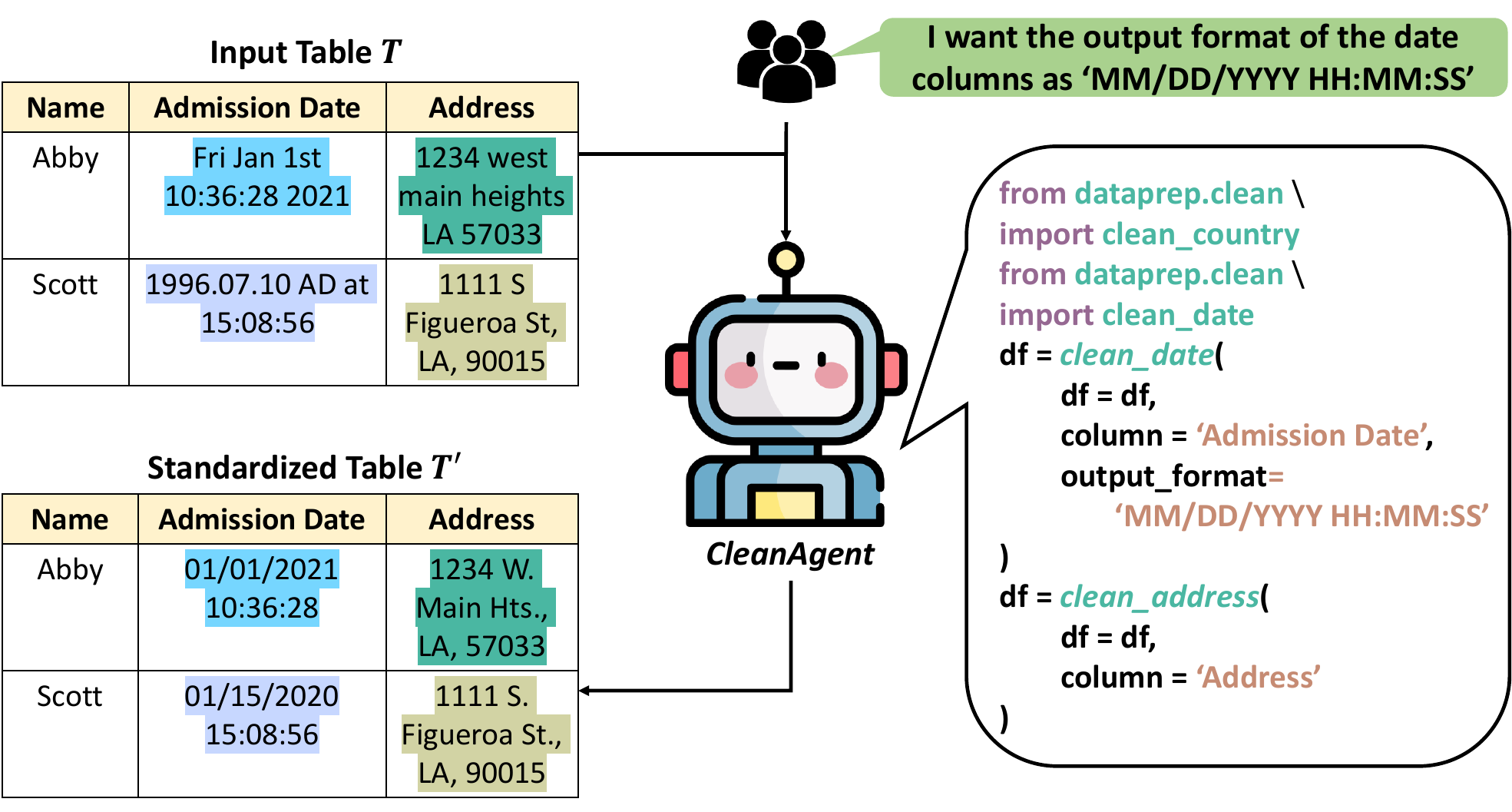}
\vspace{-1em}
\caption{An example of automatic data standardization process with CleanAgent.} 
\label{fig:example} 
\vspace{-1em}
\end{figure}

\textbf{Example 1.}
\textit{We illustrate the data standardization task in Figure~\ref{fig:example}. Given the input table $T$, it is obvious that data in the ``Admission Date'' column and the ``Address'' column are in different formats, and the data in the cells of the ``Admission Date'' column includes two different date formats. The goal of data standardization is to unify the data format in each column in $T$, to get the standardized table $T'$ satisfying the data scientist's requirements. In Figure~\ref{fig:example}, the data scientist inputs their requirement to standardize ``Admission Date'' with the ``MM/DD/YYYY HH:MM:SS'' format. In the resulting $T'$, data in the cells of the ``Admission Date'' column follows only one date format, i.e., the ``MM/DD/YYYY HH:MM:SS'' format.}


Previously, data scientists heavily relied on libraries such as Pandas~\cite{pandas} for data standardization tasks. Even though Pandas is a powerful tool, achieving data standardization often requires writing hundreds or thousands of lines of code. The standardization process for a single column involves identifying the column type, applying intricate methods such as regular expressions to each cell for validation, and converting each cell into desired formats. Moreover, a table may contain multiple columns, each possibly of a different type, requiring bespoke standardization code for each column type. 


\textbf{Example 2.} \textit{Still considering the data standardization task in Figure~\ref{fig:example}. For standardizing ``Admission Date'' and ``Address'', data scientists need to write the datetime standardization code for ``Admission Date'' and address standardization code for ``Address'' using regex. An example standardization code for ``Address'' is shown as follows}.

\begin{lstlisting}[language=Python]
def standardize_address(addr):
    # Extract street number and street name
    street = pd.Series(addr).str.extract(r'(\d+ [^,]+)').squeeze()
    # Extract state name
    state = "LA"
    # Extract zipcode
    zipcode = pd.Series(addr).str.extract(r'(\d{5})').squeeze()
    # Output standardized address
    return f"{street}, {state}, {zipcode}"
\end{lstlisting}

\textit{If the input table $T$ has other column types such as email and IP addresses, data scientists also need to write standardization code tailored for the new types, which is time-consuming. }

Recently, the emerging LLMs have shown the potential to revolutionize this process. By leveraging their natural language understanding and code generation ability, these models could significantly aid data scientists by autonomously generating standardization code in response to conversational prompts. However, this method still necessitates detailed prompt crafting and often involves multi-turn dialogues~\cite{chatpipe} for different column types in the table one by one, which limits the efficiency and practicality of adopting LLMs in the standardization process.

To overcome these limitations, our key idea is to introduce a Python library involving declarative and unified APIs specifically designed for standardizing different column types. This idea lowers the burden of the LLM, as it now only needs to convert natural language (NL) instructions into succinct, declarative API calls instead of lengthy, procedural code. Such an approach simplifies the LLM's code generation process for data standardization, requiring just a few lines of code.

The pursuit of simplicity, however, introduces two primary challenges. The first challenge (\textbf{\textit{C1}}) is the design of the declarative and unified APIs for data standardization, ensuring it can effectively reduce the intricacies involved in standardizing specific column types (ideally one line of code per column type). The second challenge (\textbf{\textit{C2}}) centers on optimizing the interaction between data scientists and LLMs. Our goal is to minimize human involvement, ideally allowing data scientists to input their standardization requirements in one instance, thereby enabling an autonomous and hands-off data standardization process. 

\textbf{\textit{To solve C1}}, we propose the type-specific \texttt{Clean} module in the
\href{https://github.com/sfu-db/dataprep}{\textcolor{blue}{Dataprep Library}}, named \href{https://github.com/sfu-db/dataprep/tree/develop/dataprep/clean}{\textcolor{blue}{\texttt{Dataprep.Clean}}}.
By observing and summarizing the common steps of data standardization for specific column types, we design unified APIs \texttt{clean\_type(df, column\_name, target\_format)}, where the \texttt{type} represents the desired standardization type, such as date, address, and phone, etc. These unified APIs offer enhanced expressiveness compared to raw Pandas code, reducing the complexity of standardizing specific column types and allowing one to standardize a column with only one line of code. 


\textbf{\textit{To solve C2}}, we propose the \cleanagent framework which automates data standardization with \texttt{Dataprep.Clean} and LLM-based Agents
~\cite{AgentSurvey2, AutoGen}. Once users have entered their final goals, the LLM-based Agents can free their hands, autonomously generate reasoning steps, and execute particular tasks. 
Data scientists only need to 
input the table being standardized and their requirements, \cleanagent will complete the data standardization process automatically with three steps: annotating the type of each column, generating concise Python code for standardization, and executing the generated Python code.

\textbf{Example 3.}
\textit{Continuing with Example 1. Given an input table $T$ which needs to be standardized and the data scientists' requirements, the \cleanagent first recognizes that the ``Admission Date'' column belongs to the date type, and the ``Address'' column belongs to the address type. According to the column-type annotation results, the \cleanagent generates and executes Python code for standardization by calling the ``clean\_date'' and ``clean\_address'' functions, then returns the standardized table $T'$.}

We also built a web interface for \cleanagent. It allows the users to choose sample data and communicate with \cleanagent for standardization. We provide the demonstration video, which can be found on \href{https://youtu.be/fSYXVM6qeqM}{\textcolor{blue}{Youtube}}.

 To summarize, we make the following contributions: (1) We propose \texttt{Dataprep.Clean}, an open-sourced library for reducing the complexity of implementing data format standardization with type-specific standardization functions. (2) We propose \cleanagent, which automates the data standardization process by combining both the advantages of \texttt{Dataprep.Clean} and LLM-based Agents. (3) We deploy \cleanagent as a web application with a user-friendly interface and demonstrate its utility. We also open-sourced the implementation of \cleanagent on \href{https://github.com/sfu-db/CleanAgent}{\textcolor{blue}{Github}}.
\vspace{-.5em}
\section{Type-Specific Standardization API Design}
In this section, we first describe the common steps of data standardization. Then, we introduce the type-specific API design of \texttt{Dataprep.Clean}.

\noindent\textbf{Common Steps of Data Standardization.}
Inspired by the steps of how human users standardize data cells, we identify three common steps of data standardization. We take the \texttt{datetime} column type as an example to illustrate these steps.

Assume a data scientist is dealing with an \texttt{datetime} column including two records \textit{"Thu Sep 25 10:36:28 2003"} and \textit{"1996.07.10 AD at 15:08:56"}. The data scientist wants to unify the messy column into a target format \textit{"YYYY-MM-DD hh:mm:ss"}.


\underline{\textit{(1) Split.}}
In the beginning, the data scientist needs to split the datetime string into several single parts, which include one kind of specific information. In our example, the data scientist can get several tokens \texttt{\{'Thu','Sep','25','10','36','28','2003'\}} from the first record by using space and colon as separators. Each different type has its splitting strategy, which may not always be splitting the string into tokens. For example, the data scientist will split the email string into the \texttt{username} part and the \texttt{domain} part.

\underline{\textit{(2) Validate.}} Standardization can only be performed on valid inputs. Thus, the second step should be validation. For example, if the string ``little cat'' is an instance of the \texttt{datetime} column, this string is invalid, and the data scientist will transform it to a default value like \texttt{NaN}. Intuitively, a valid string indicates that each part of this string after splitting is valid. 
Usually, the data scientist will recognize and validate each part by their domain knowledge, some corpus or some rules. If every split part is valid, the string is also valid. For instance, the token \texttt{'Sep'} can be recognized as a valid representation of a month, and \texttt{'2003'} can be recognized as a valid year.

\underline{\textit{(3) Transform.}} The last step of standardization is to transform each split part and combine them into the target format. In our example, because the target format is \textit{"YYYY-MM-DD hh:mm:ss"}, the month \texttt{Sep} is transformed into number \texttt{09} and recombined with other parts to the target \texttt{"2003-09-25 10:36:28"}.  


\noindent \textbf{The Design of Unified APIs.} The goal of our API design is to enable data scientists to complete all the common steps of standardizing one column with a single function call. Simplicity and consistency are considered the principles of API design. The observation of the common steps of data standardization brings the type-specific API design idea.  More specifically, we design the API to be in the following form:
\begin{align*}
    \textbf{{clean\_}\textit{type}}
    \texttt{(df, column\_name, target\_format)}
\end{align*}

\begin{figure}[t] 
\vspace{-1.5em}
\centering 
\includegraphics[width=1.0\linewidth]{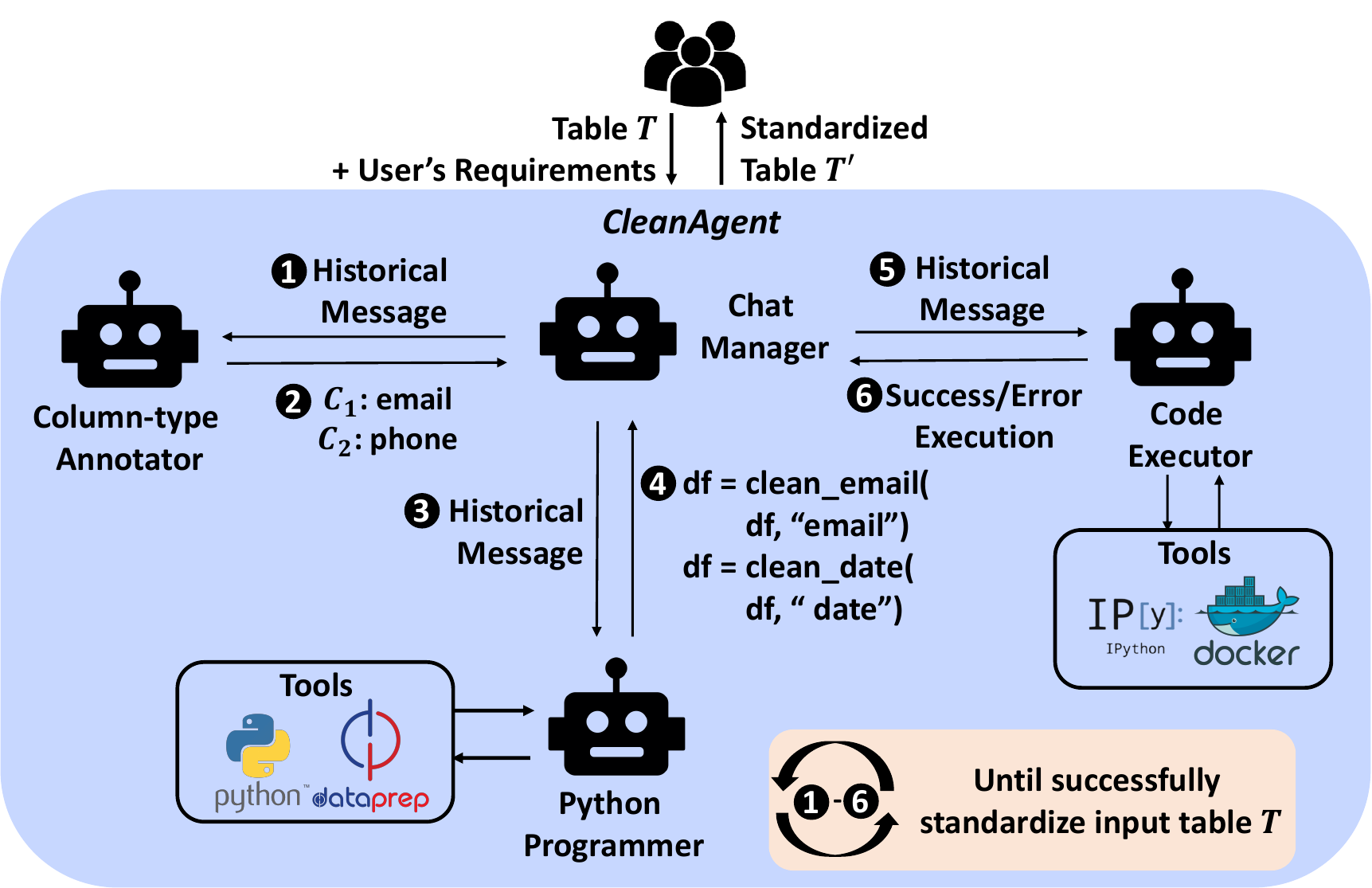}
\vspace{-1em}
\caption{The Workflow of CleanAgent.} 
\label{fig:clean_agent_framework} 
\end{figure}

where \texttt{clean\_type} is the function name, \texttt{type} represents the type of the current column. The first argument \texttt{df} represents the input DataFrame, the second argument 
\texttt{column\_name} is the column being standardized, and the third argument \texttt{target\_format} is the target standardization format users specified. Our API design is flexible and extensible, which makes it convenient for users to add their standardization functions for new data types. Currently, we have 142 standardization functions in \texttt{Dataprep.Clean}, each handles one data type. These functions serve to demonstrate the value of a more declarative approach, illustrating that building declarative data standardization tools for LLMs is not only feasible but essential, motivating the community to develop even more advanced tools.

\vspace{-.5em}
\section{CleanAgent Workflow}
In this section, we first introduce the basic structure of LLM-based agents. Then, we describe the \cleanagent workflow constructed by four agents. The automatic data standardization process can be completed by the cooperation of the four agents in \cleanagent. 

\noindent \textbf{Basic Structure of LLM-based Agents.}
According to the previous surveys on LLM-based Agent~\cite{AgentSurvey2}, an LLM-based agent includes four main components: (1) a backbone LLM used to generate replies for input prompts, (2) a memory used to store historical conversation messages, (3) a system message defining the role of the agent, and (4) a set of external tools which can be called by the LLM-based agent to complete specific tasks, such as web searching, code execution, etc.

\noindent \textbf{Detailed Workflow.}
The detailed workflow of \cleanagent is shown in Figure \ref{fig:clean_agent_framework}. The \cleanagent is composed of four agents, including a \textit{Chat Manager}, a \textit{Column-type Annotator}, a \textit{Python Programmer}, and a \textit{Code Executor}. 
They can communicate with each other and automatically complete the data standardization process by cooperation. Each agent has its own memory to store the historical conversational messages between it and other agents. Note that the memory of the \textit{Chat Manager} is uniquely comprehensive, encompassing the entire historical conversational messages from all agents within the \cleanagent system. This extensive memory enables every agent in the \cleanagent to generate responses that are informed by the complete historical messages.

The input of \cleanagent includes a table $T$ that needs to be standardized. Data scientists can also input extra requirements such as ``the format of the date type column should be MM/DD/YYYY''. 
By receiving the input table and data scientists' extra requirements, \cleanagent stores this information into the \textit{Chat Manager's} memory and then completes the data standardization process. The \textit{Chat Manager} delivers messages in its memory to the \textit{Column-type Annotator}(\ding{172} in Figure~\ref{fig:clean_agent_framework}). 
Then, \textit{The Column-type Annotator} receives the table information and leverages an LLM to annotate the type of each column in the input table. If the \textit{The Column-type Annotator} cannot figure out the specific type of one column, the \textit{Column-type Annotator} outputs ``I do not know''. The annotation result is returned to the \textit{Chat Manager} and stored in the \textit{Chat Manager}'s memory (\ding{173} in Figure~\ref{fig:clean_agent_framework}). 

\begin{figure*}[t] 
\vspace{-1.5em}
\centering 
\includegraphics[width=0.8\linewidth]{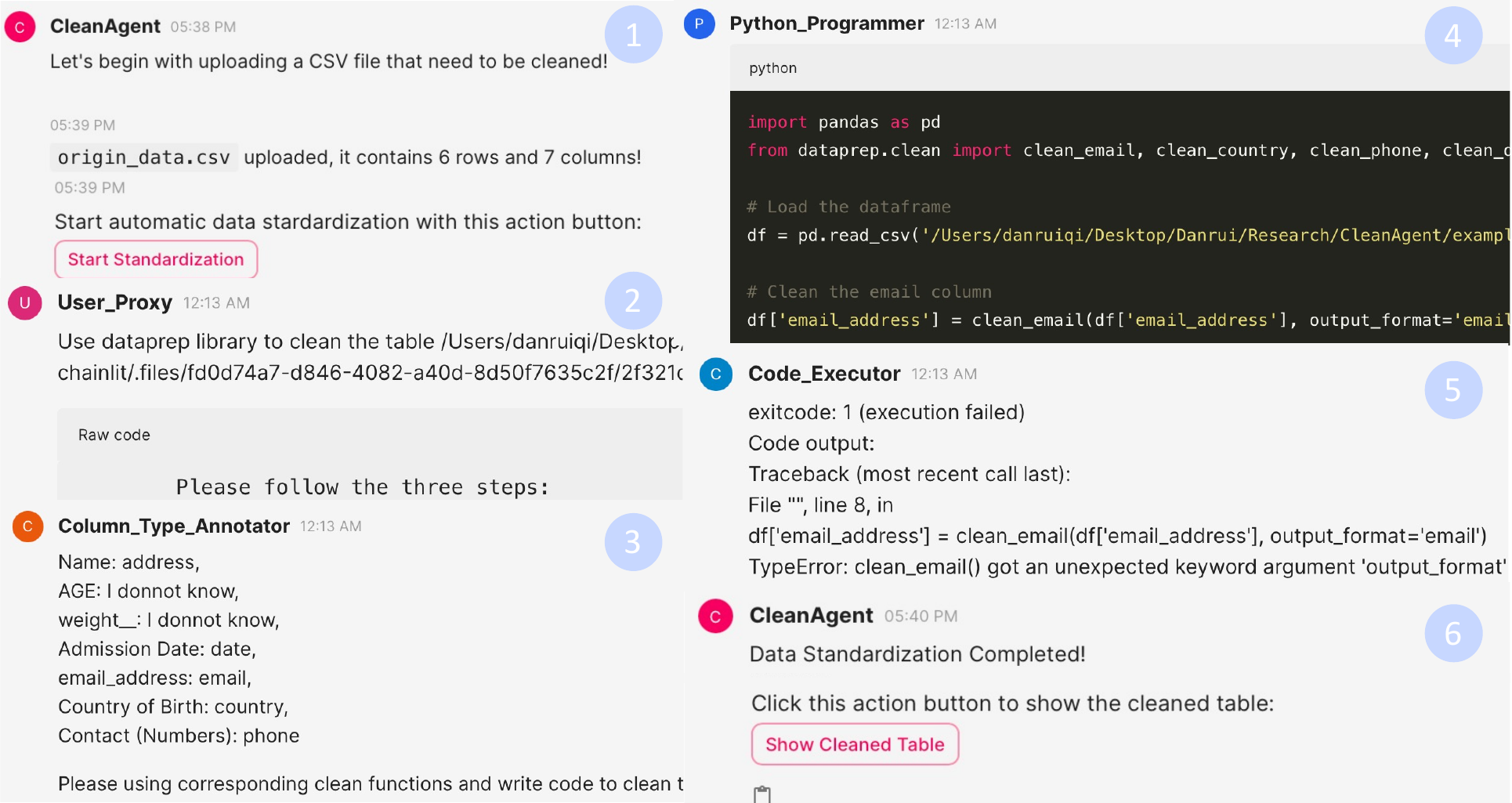}
\vspace{-1em}
\caption{User interface of CleanAgent.} 
\label{fig:demo} 
\end{figure*} 

Thirdly, the \textit{Python Programmer} receives historical messages from the \textit{Chat Manager} including the column-type annotation results (\ding{174} in Figure~\ref{fig:clean_agent_framework}), picks up the corresponding clean functions, and generates Python code for the data standardization process. The generated Python code is also returned to the \textit{Chat Manager} and stored in the \textit{Chat Manager}'s memory (\ding{175} in Figure~\ref{fig:clean_agent_framework}). 
Finally, the \textit{Code Executor} receives historical messages from the \textit{Chat Manager} including the column-type annotation results and the generated Python code (\ding{176} in Figure~\ref{fig:clean_agent_framework}), then executes the generated Python code. If the generated code executes without errors, the standardized table $T'$ is returned; otherwise, the error message is returned to the \textit{Chat Manager} and stored in its memory (\ding{177} in Figure~\ref{fig:clean_agent_framework}). Then, \cleanagent will retry the whole workflow until it can complete the data standardization process successfully.







\section{Experiments}
\noindent \textbf{Dataset.}
In our experiment, we employ the \textit{\textbf{Flights}} dataset from~\cite{holoclean}, as it contains highly irregular datetime formats across four attributes: \texttt{scheduled\_dept, actual\_dept, scheduled\_arrival, and actual\_arrival}. The datetime values in these columns exhibit a wide variety of inconsistent formats, such as \texttt{"2011-12-08 3:50:00 PM", "2:30pDec 27", "06:45 AM Sun 25-Dec-2011"}, etc. This makes the dataset particularly suitable for evaluating the standardization capabilities of different systems.

 \noindent\textbf{Baselines.} We compare \cleanagent with the following two baselines: \underline{\textit{(1) GPT-4o + Prompting.}} Data standardization code can be directly generated by prompting powerful chat models such as GPT-4o. \noindent \underline{(2) \textit{Cocoon}~\cite{Cocoon}}. Cocoon is a one-shot data cleaning system that decomposes complex cleaning tasks into manageable components within a workflow designed to mimic human cleaning processes, leveraging large language models. It supports a variety of data cleaning tasks, including missing value imputation, outlier detection, and functional dependency violation. In this paper, however, we focus on evaluating Cocoon's ability for data standardization.

Note that there are other LLM-based data cleaning approaches, such as \textit{RetClean~\cite{Retclean}}. However, \textit{RetClean} primarily adopts a retrieval-based strategy such as RAG to enhance the ability of LLMs for data cleaning, which supplements the LLM with user-provided data sources. This paradigm is not suitable for our scenario.

\noindent \textbf{Ground Truth Generation.} We find that GPT-4o can reliably convert individual datetime strings into a target format (e.g., YYYY-MM-DD HH:MM:SS). Thus, we use it to generate cell-level ground truth values and compile them into a complete table.
 
\noindent \textbf{Metrics.} We use the average cell-level matching rate across all columns as our evaluation metric. For a given table $T$, the cell-level matching rate is computed as:

\begin{align}
d(T_{\text{clean}}, T_{\text{gt}}) = \frac{\sum_{i = 1}^{m}\sum_{j = 1}^{n} \mathbf{1}(T_{clean_{ij}} = T_{gt_{ij}})}{m}
\end{align}

\noindent where $\mathbf{1}[\cdot]$ is the indicator function, and $T_{\text{clean}}$ and $T_{\text{gt}}$ denote the standardized and ground truth tables, respectively.

\noindent \textbf{Implementation.} \noindent \textbf{Implementation.} \cleanagent is implemented in Python 3.10.6. Cocoon is run using its official Colab notebook\footnote{\url{https://colab.research.google.com/github/Cocoon-Data-Transformation/cocoon/blob/main/demo/Cocoon_Stage_Demo.ipynb}} from the GitHub repo\footnote{\url{https://cocoon-data-transformation.github.io/page/clean}}. All methods use the \texttt{gpt-4o-2024-08-06} model. Experiments are conducted on a MacBook Pro with an M1 chip, 16GB RAM, running macOS Sequoia 15.5.

\noindent \textbf{Results.} Table~\ref{tab:experiments} presents the comparison of different systems in terms of cell-level matching rate and latency. \cleanagent achieves a 42.5\% cell-level matching rate, approximately \textbf{2$\times$} higher than that of GPT-4o and Cocoon. These results demonstrate that \cleanagent’s type-specific standardization API enhances the LLM’s ability to generate more precise and concise standardization code. In addition to higher accuracy, \cleanagent also exhibits lower latency compared to Cocoon. This is because Cocoon generates a one-shot SQL query for all columns without the ability to target specific ones, leading to unnecessary overhead.

\begin{table}[htbp]
\caption{Data standardization performance by comparing different systems.}

\label{tab:experiments}
\centering
\begin{tabular}{lcc}
\hline
\textbf{System}     & \makecell{\textbf{Cell-Level} \\ \textbf{Matching Rate(\%)}} & \textbf{Latency (s)} \\ \hline
\textbf{GPT-4o}     & 22.0                        & 19.76                \\
\textbf{Cocoon}     & 21.5                        &   636.62                   \\
\textbf{CleanAgent} & \textbf{42.5}                        & 29.57                \\ \hline
\end{tabular}
\end{table}

\section{User Interface of CleanAgent}

We developed a web-based user interface for \cleanagent, allowing users to simply upload their tables without performing any operations. The system then automatically returns the standardized results of their data.

Figure~\ref{fig:demo} shows the user interface of \cleanagent. As area \ding{172} shows, users must first upload a CSV file that needs to be cleaned. Then \cleanagent shows the basic information of the uploaded file (number of rows and number of columns). If the users can click the ``Start Standardization'' button to start the data standardization process by want \cleanagent.

After clicking the ``Start Standardization'' button, as area \ding{173} shows, the \texttt{User\_Proxy} generates three detailed steps to complete the data standardization task. Firstly, the \texttt{Column-type Annotator} receives messages from the \texttt{Chat Manager}, annotates and outputs the type of each column, as area \ding{174} shows. Then, the \texttt{Python Programmer} picks up standardization functions from Dataprep.Clean based on the type of each column, and write proper Python code using the standardization functions, as area \ding{175} shows. Thirdly, the \texttt{Code Executor} executes the Python code by the \texttt{Python Programmer} and collects the execution messages, as area \ding{176} shows. If the \texttt{Code Executor} gets an error message when executing generated Python code, the error message is sent to the \textit{Chat Manager} and becomes part of the prompt of the next try. If the \texttt{Code Executor} gets the message of successful execution, \cleanagent will report that the data standardization is completed, as area \ding{177} shows. Moreover, users can click the ``Show Cleaned Table'' button to check whether the standardized table matches their requirements. If so, users can download the standardized table directly. Otherwise, users can input their extra requirements with natural language, and \cleanagent will start a new data standardization process accordingly.

\section{Conclusion}
In this paper, we proposed \cleanagent to automate the data standardization process with \texttt{Dataprep.Clean} and LLM-based Agents. We implemented \cleanagent as a web application to visualize the conversations among agents.
Other tasks in the data science life cycle, such as data cleaning and data visualization, can also be completed by LLM-based agents~\cite{DB-GPT}. 
In the future, it is promising that the data science life cycle can be automatically planned and completed by LLM-based agents' cooperation.

\bibliographystyle{ACM-Reference-Format}
\bibliography{references}

\begin{appendix}
\section{Prompts of Component in CleanAgent}

\begin{tcolorbox}[colback=cyan!5!white,colframe=cyan!75!black, title=System Message of Chat Manager]
Use dataprep library to clean the table \{path\}.

Please follow the three steps:
\begin{enumerate}
    \item Use column annotator to annotate the type of each column within the five types: \{candidate\_column\_types\}. 
    \item Pick up corresponding clean functions and write code to clean the column.
    \item store the cleaned dataframe as csv file named as "cleaned\_data.csv"
\end{enumerate}

\end{tcolorbox}

\begin{tcolorbox}[colback=cyan!5!white,colframe=cyan!75!black, title=System Message of Column Annotator]
You are an expert column type annotator.

Please solve the column type annotation task following the instruction. Please ALWAYS show the column annotation result!!! Please ONLY return the column annotation result adding a sentence "Please using corresponding clean functions and write code to clean the column"!!! 

Classify the columns of a given table with only one of the following classes that are seperated with comma: \{candidate\_column\_types\}.

\begin{enumerate}
    \item Look at the input given to you and make a table out of it.
    \item  Look at the cell values in detail.
    \item For each column, select a class that best represents the meaning of all cells in the column.
    \item Answer with the selected class for each columns with the format **columnName: class**. If you cannot confidently classify a column based on the provided data, output "I do not know" for that column.
\end{enumerate}

NOTE THAT You MUST provide exactly one classification for EVERY column — no column should be left unclassified.

Sample rows of the given table is shown as follows: \{df\}.
\end{tcolorbox}

\begin{tcolorbox}[colback=cyan!5!white,colframe=cyan!75!black, title=System Message of Python Code Generator]
You are a senior Python engineer who is responsible for writing Python code to clean the input DataFrame. 

You can use the following libraries: pandas, numpy, re, datetime, dataprep, and any other libraries you want. Note that the Dataprep library takes the first priority.
                                
The Dataprep library is used to standardize the data. You can find the documentation of Dataprep library here: https://sfu-db.github.io/dataprep/.

Please only output the code.
\end{tcolorbox}

\begin{tcolorbox}[colback=cyan!5!white,colframe=cyan!75!black, title=System Message of Python Code Executor]
You are a Python code executor that executes the code written by the engineer and reports the result.
\end{tcolorbox}

\section{Detailed Experiment Settings}
\subsection{Prompt of GPT-4o Baseline}
\begin{tcolorbox}[colback=cyan!5!white,colframe=cyan!75!black, title=System Message of Chat Manager]
You are an expert data standardizer.

Task: Given a CSV file **raw.csv** in the current working directory, do two things:

1. Column typing  

Inspect the data and output one best-fit type for each column, line by line in the form:  

columnName: class

2. Generate Python script  

After a blank line, provide a single Python script (inside ```python fences) that:

- reads raw.csv

- standardizes every column WITHOUT USING ANY Python libraries

- no additional explanations.

- please notice the python code, **please not using any libraries** such as**datetime, parse, colorsys, pandas**. Only the original way and regex can be used.

- if a cell cannot be recognized according to the column’s target format, return `NaN`.

- formatting rules for column types: 
\begin{enumerate}

  \item date → yyyy-mm-dd hh:mm:ss  
  \item address → Apt apartment\_number, house\_number, street\_name, city, state\_abbreviation, country, zipcode (skip any missing part silently)  
  \item phone\_number → E.164 format  
  \item location → (lat,lon)  
  \item ip → plain IP without subnet mask  
  \item url → JSON object with keys:
  
     \{
     
       'scheme': 'http',  
       
       'host': 'www.example.com',  
       
       'url\_clean': 'http://www.example.com/path',  
       'queries': \{
       
         'key1': 'value1',
         
         'key2': 'value2'
         
       \}
       
    \}  
  \item duration → hh:mm:ss  
  \item temperatures → Celsius format, e.g., 23℃
  \item colors → hexadecimal, e.g., \#a1b2c3  
 \item names → "firstname lastname"  
     - If format is "lastname, firstname", convert it  
     - If already "firstname lastname", keep it unchanged   
\end{enumerate}

Writes cleaned\_data.csv in the same directory  

The script must be runnable with `python script.py` in a standard Python environment (pandas \& common pip packages installed)

Return only the column typing and script. No additional explanations.

Sample rows of the given table is shown as follows: \{df\}
\end{tcolorbox}

\subsection{Detailed GPT Settings}
For \cleanagent, we use GPT-4o with a temperature of 0, a timeout of 60 seconds, and a cache seed of 42. For Cocoon~\cite{Cocoon}, we follow the default setting and set the temperature to 1. For the GPT-4o baseline, the temperature is also set to 0 for consistency with \cleanagent.
\end{appendix}

\end{document}